# Introducing Representations of Facial Affect in Automated Multimodal Deception Detection

Leena Mathur
University of Southern California
Department of Computer Science
Los Angeles, CA, USA
lmathur@usc.edu

Maja J Matarić
University of Southern California
Department of Computer Science
Los Angeles, CA, USA
mataric@usc.edu

## Abstract

Automated deception detection systems can enhance health, justice, and security in society by helping humans detect deceivers in high-stakes situations across medical and legal domains, among others. Existing machine learning approaches for deception detection have not leveraged dimensional representations of facial affect: valence and arousal. This paper presents a novel analysis of the discriminative power of facial affect for automated deception detection, along with interpretable features from visual, vocal, and verbal modalities. We used a video dataset of people communicating truthfully or deceptively in real-world, high-stakes courtroom situations. We leveraged recent advances in automated emotion recognition in-the-wild by implementing a state-of-the-art deep neural network trained on the Aff-Wild database to extract continuous representations of facial valence and facial arousal from speakers. We experimented with unimodal Support Vector Machines (SVM) and SVM-based multimodal fusion methods to identify effective features, modalities, and modeling approaches for detecting deception. *Unimodal models trained on facial affect achieved an AUC of 80%, and facial affect contributed towards the highest-performing multimodal approach (adaptive boosting) that achieved an AUC of 91% when tested on speakers who were not part of training sets. This approach achieved a higher AUC than existing automated machine learning approaches that used interpretable visual, vocal, and verbal features to detect deception in this dataset, but did not use facial affect.* Across all videos, deceptive and truthful speakers exhibited significant differences in facial valence and facial arousal, contributing computational support to existing psychological theories on relationships between affect and deception. The demonstrated importance of facial affect in our models informs and motivates the future development of automated, affect-aware machine learning approaches for modeling and detecting deception and other social behaviors in-the-wild.

## 1 Introduction

Advances in *social signal processing* and *multimodal machine learning* are enabling the development of machines that can automatically detect and recognize human behaviors, including the social behavior of *deception* [5, 60]. Deception involves the intentional communication of false or misleading information and is often categorized as occurring in either *high-stakes* or *low-stakes* situations [23], with deceivers in high-stakes contexts facing substantial consequences if their deception is discovered. To create a more healthy, secure, and just society, psychologists, social workers, government agencies, and law enforcement groups have fostered a growing interest in detecting deception in high-stakes situations such as therapist-client communications (e.g., during self-disclosure of abuse or mental health issues) [12], courtroom testimonies [10], and forensic investigations [50]. Humans, even experts trained in lie detection, experience difficulty in accurately and precisely detecting deception in social situations. Meta-analysis of human deception detection abilities finds them close to chance level [8], motivating the development of automated approaches that can support humans in this challenging task.

Recent approaches for automated deception detection have focused on developing machine learning models that exploit discriminative patterns in behavioral cues to distinguish deceptive and truthful communication. Video-based deception detection is a current priority in deception research, because behavioral cues can be extracted from videos in a cheaper, faster, and *non-invasive* manner [10], which is preferable to *invasive* approaches that extract cues through devices attached to human bodies (e.g., polygraphs). Another priority is leveraging more *interpretable* features and machine learning approaches for deception detection [31], since interpretability is an important aspect of automated systems deployed to help humans in real-world, high-stakes situations. Current methods for video-based deception detection have combined and used interpretable features from human behavioral cues in visual (e.g., facial movements), vocal (e.g., pitch), verbal (e.g., word choice), and physiological (e.g., thermal imaging) modalities [10].

Existing machine learning approaches for automated deception detection in videos have not included continuous representations of affect, neurophysiological states (e.g., "pleasure") that can be components of emotions or moods [53]. Affect is a fundamental aspect of human communication that interacts with cognition to guide social behaviors in human-human [22, 57], human-computer [41], and human-robot [26] interactions. Since the face is a key channel through which affect is expressed [3], facial affect is a promising modality to consider when developing automated systems for detecting social behaviors such as deception. The 2-D *valence* and *arousal* space [53] is a standard model for representing the subtle and complex nature of affective states by capturing how pleasant or unpleasant each state is (valence) and how passive or active each state is (arousal) [27]. Psychological studies of affect and deception have theorized that deceivers in high-stakes situations are likely to exhibit affective states with lower valence and higher arousal, relative to truthful speakers, expressed through temporal patterns of involuntary nonverbal cues [18, 65]. *We hypothesized that temporal patterns in representations of facial valence and facial arousal could be effectively leveraged to automatically detect deception in real-world, high-stakes situations.*

Inspired by other high-stakes deception detection studies [4, 47, 48, 51, 61], we conducted experiments with a real-world video dataset of people speaking truthfully or deceptively in courtrooms [47]. We leveraged recent advances in affective computing in-the-wild to automatically extract continuous representations of facial valence and facial arousal from speakers in the videos. Since deception is communicated through verbal and nonverbal cues, and affect can influence these cues [22], our goal was to analyze the discriminative power of facial affect for automated deception detection, along with interpretable features from human verbal and nonverbal communication. We experimented with unimodal and multimodal SVM-based machine learning approaches that used *facial affect* (valence, arousal), *visual* (face, head, eye movements), *vocal* (spectral, cepstral, prosodic, voice quality), and *verbal* (psychometric) cues to detect deception. This paper, to the best of our knowledge, presents the first use and evaluation of continuously-valued facial affect for automated deception detection in videos.

Our results demonstrate the discriminative power of facial affect for automated deception detection in videos. Unimodal models trained on facial affect features achieved an AUC of 80%, and facial affect contributed towards the highest performing multimodal approach, which obtained an AUC of 91% through adaptive boosting trained on facial affect, visual, and vocal features. This multimodal approach outperformed existing automated machine learning approaches that have used interpretable visual, vocal, and verbal features with this dataset, but have not used facial affect [4, 47, 51, 61]. Across all videos, deceptive and truthful speakers exhibited significant differences in mean, median, and standard deviation of facial valence and facial arousal, contributing computational support to psychological theories on affect and deception. Our results aim to inform and motivate the development of affect-aware approaches for automated deception detection in-the-wild.

This work makes the following contributions:
- Development of a novel, automated approach for video-based deception detection that uses continuous representations of facial affect, along with interpretable visual, vocal, and verbal features from human communication.
- Evaluation of unimodal and multimodal SVM-based machine learning approaches to identify effective features, modalities, combinations of modalities, and modeling approaches for affect-aware automated deception detection.
- Novel analysis of the computational relationship between facial affect and deception observed in this real-world high-stakes dataset, contributing support to psychological theories on relationships between affect and deceptive behavior.

Section 2 of this paper describes the related works that inform this research. Section 3 details the experimental methodology. Section 4 presents an analysis of relationships between facial affect and deception in the dataset. Section 5 contains the results and discussion of the unimodal and multimodal deception detection models. Section 6 concludes the paper.

## 2 Related Works

This section provides an overview of key areas that contribute to this research: (1) deception detection, with a focus on video-based deception detection in high-stakes situations, (2) psychological theories on the relationship between affect and deception, and (3) automated facial affect recognition from videos in-the-wild.

### 2.1 Deception Detection

Early systems for detecting deception relied on polygraph tests, an *invasive* approach that requires attaching devices to the human body to collect physiological data (e.g., skin conductance) [10] that human experts would interpret in order to detect deception. The deception research community has recently shifted focus to more reliable and *non-invasive* approaches that remotely extract and use verbal and nonverbal behavioral cues to detect deception [10, 31, 51, 61]. Video-based deception detection approaches are of particular interest because they can remotely sense behavioral cues and can be deployed at a large-scale in real-world situations [10].

Decades of research in psychology, forensics, and deception detection have documented verbal and nonverbal behavioral cues indicative of deceptive communication. These insights can and have been leveraged when choosing features to use in automated deception detection systems. Visual cues such as the frequency and duration of eye blinks [7, 24, 42, 45], dilation of pupils [16, 40], head movements [39], and facial muscle movements [28, 49] have been found to distinguish between deceptive and truthful behavior. Vocal cues can be indicative of deception, with deceptive speakers tending to speak with higher and more varied pitch [14, 65], shorter utterances, and less fluency [52, 56] than truthful speakers. Representations of vocal affect (valence and arousal) have been used to detect deceptive speech by using models trained on emotional speech databases to generate emotional features used in deception classification [2]. Deception also correlates with verbal attributes of speech, with deceivers tending to communicate with less cognitive complexity, fewer self-references, and more words indicative of negative emotions [44, 64]. To the best of our knowledge, representations of facial affect (valence and arousal) have not previously been used for automated deception detection.

Recent research in video-based deception detection focuses on detecting deception in real-world, high-stakes situations. Pérez-Rosas et al. [47] developed the first publicly-available, high-stakes deception video dataset (60 truthful videos, 61 deceptive videos, ~28 seconds per video) of people speaking in real-world courtrooms. This dataset is the current benchmark for multimodal high-stakes deception detection in videos and was used for the work in this paper. Benchmark accuracies using manually-annotated hand gestures, facial displays, and head movements, as well as automatically-extracted transcript unigrams and bigrams, range from 60-75% with decision-trees (DT) [47]. While some approaches for deception detection in this dataset have used deep learning [15, 25, 35], deep learning approaches for deception detection have been considered inadvisable in this dataset due to its small size [15, 31, 61]. In this paper, we evaluate the effectiveness of our models by comparing our results to those of existing automated machine learning approaches (performances summarized in Table 1) that have used classifiers such as SVM, Decision Trees, Random Forests, Logistic Regression, and AdaBoost to detect deception in this dataset [4, 29, 47, 51, 61]. We examined papers that have cited the Pérez-Rosas et al. dataset [47] in order to compile the performances of existing automated approaches for deception detection. The highest-performing automated approach [61] achieved an AUC of 87.7% by using the following interpretable visual, vocal, and verbal features in late fusion with a Linear SVM: (1) head, face, and eye movements, (2) MFCC coefficients, and (3) transcript GloVe word embeddings.

Table 1: Highest-performing machine learning approaches from existing automated video-based deception detection with the Pérez-Rosas et al. [47] courtroom dataset.

| Modality | Model | AUC | ACC |
| --- | --- | --- | --- |
| Visual + Verbal [47][†] | DT early fusion | — | 75% |
| Visual + Vocal + Verbal [29] | SVM early fusion | — | 79% |
| Visual + Vocal + Verbal [61] | SVM late fusion | 88% | — |
| Visual [4] | SVM unimodal | — | 77% |
| Visual, Vocal (tie) [51] | SVM unimodal | 70% | — |

[†]Benchmark (visual features not automatically extracted)

## 2.2 Affect and Deception

Psychologists have developed theories regarding the relationship between affect and deception. Ekman and Friesen [18] proposed the *leakage hypothesis*, which posits that deceivers experience and exhibit negative and aroused physiological states that involuntarily "leak" into the movements of their faces and bodies. Zuckerman et al. [65] developed the prominent *four-factor theory* of deception, which attempts to explain this involuntary "leakage" as a result of four interacting causes: (1) general physiological arousal, (2) negative affective states, (3) increased cognitive effort, and (4) unsuccessful attempts at controlling behavioral cues. Deceivers in high-stakes situations are likely to experience and exhibit affective states with lower valence and higher arousal, associated with anxiety and fear at the possibility of their deception being revealed, as well as guilt at engaging in deception [18, 37, 65]. It is worth noting that deceivers do not universally emit observable negative and aroused affective states, due to differences in their situations, reactions to situations [38], ability to conceal internal affective states, and behavioral cues when leaking internal affective states. These insights from *four-factor theory* motivated our incorporation of facial affect in models for automated high-stakes deception detection.

## 2.3 Facial Affect Recognition In-the-Wild

Since humans express emotions differently in naturalistic real-world situations compared to laboratory or controlled contexts, current research in automated facial affect recognition focuses on detecting facial affect *in-the-wild*. Kollias et al. [33, 34, 63] collected the first large video dataset of facial affect in-the-wild (Aff-Wild) that spans diverse environments, affective states, ethnicities, and recording conditions (e.g., varied poses, illuminations, and occlusions). This benchmark dataset contains 298 videos, 200 subjects, and over 30 hours of data, with facial valence and facial arousal continuously annotated between -1 and 1 at each frame by 8 experts. Kollias et al. also developed the AffWildNet [34], a benchmark and state-of-the-art deep neural network that leverages a CNN and RNN architecture to predict facial valence and facial arousal in-the-wild in this dataset. AffWildNet learns rich representations of valence and arousal and has been demonstrated as capable of generalizing its knowledge in other contexts outside of the Aff-Wild dataset [34]. Our research used AffWildNet to extract representations of facial valence and facial arousal from videos of people speaking truthfully and deceptively in courtrooms in-the-wild.

## 3 Methodology

Our approach for automated deception detection in videos contains the following three steps: (1) multimodal feature extraction, (2) feature selection, and (3) classification experiments.

### 3.1 Video Dataset

A deception dataset of people speaking truthfully or deceptively in 121 real-world trial videos [47] (60 truthful videos, 61 deceptive videos, ~28 seconds per video) was used for our experiments. This dataset was collected under unconstrained situations in-the-wild during courtroom testimonies and contains speakers of diverse genders and ethnicities. Creators of this dataset labeled videos as "truthful" or "deceptive" per testimony verified by police investigations; we used these labels as ground truth. While each video primarily focuses on one speaker, the following challenges exist in this dataset: some videos briefly switch focus among faces in the courtroom, and there are variations in video illumination, camera angles, and face sizes, as well as the occasional obstruction of faces behind objects. Similar to [51, 61], we identified a subset of videos as unusable for processing facial information, discarding those that contained a sideways profile view of the speaker's face, objects obstructing the speaker's face, or unreasonably grainy video quality. Our criteria rendered a usable dataset of 108 videos (53 truthful videos, 55 deceptive videos). There are 47 unique individuals in the subset of videos used in our experiments.

### 3.2 Multimodal Feature Extraction

To capture speakers' verbal and nonverbal behaviors from videos in the dataset, we extracted 191 interpretable features that span four channels of information: a speaker's facial affect, visual, vocal, and verbal cues. We focused on extracting interpretable features, instead of deep representations, in order to more effectively identify and analyze verbal and nonverbal behavioral cues associated with deception, similar to feature analysis conducted in [58]. The multimodal feature extraction process is visualized in Figure 1.

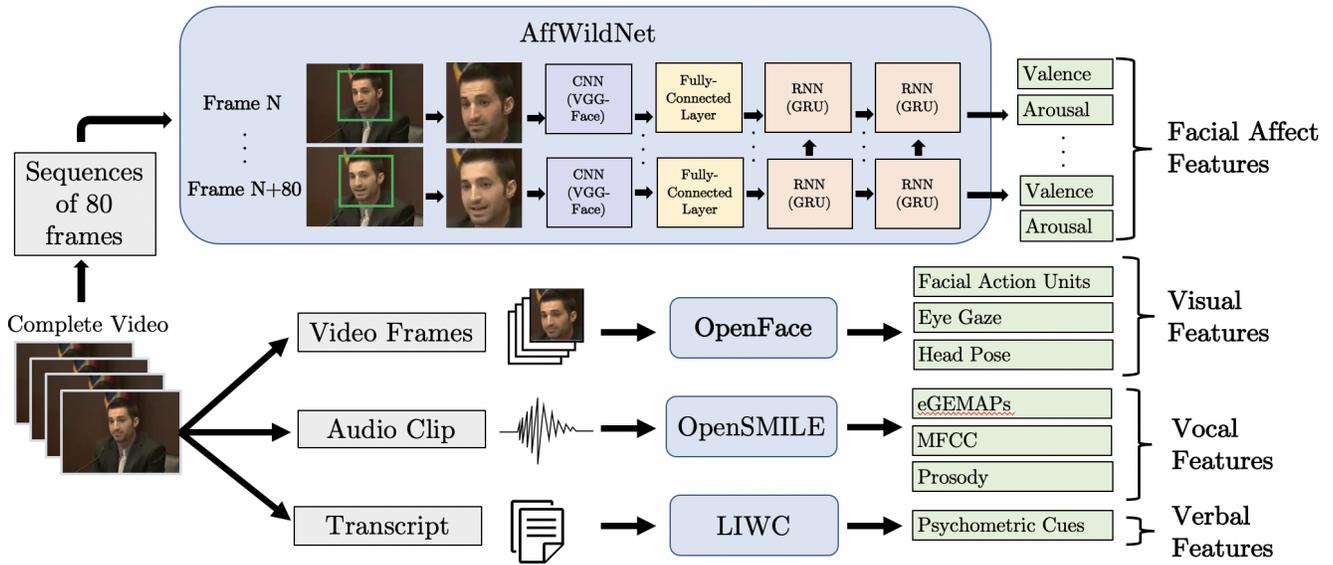

**Figure 1: Multimodal Automated Feature Extraction Process.**

*3.2.1 Facial Affect* A state-of-the-art AffWildNet model trained on the Aff-Wild database [34] was used to extract facial affect representations of valence and arousal from the videos. We implemented AffWildNet in TensorFlow [1], with model weights available from the developers. To prepare facial frames as input for AffWildNet, we tracked the face of the primary speaker from each video, extracted images of the facial bounding boxes at each frame, resized images to 96x96x3, and normalized the images' pixel intensity values between -1 and 1. Facial frames were prepared as Tensors with sequences of 80 consecutive images (per AffWildNet hyperparameters) and fed through AffWildNet to extract continuous valence and arousal predictions between -1 and 1. The graphs in Figure 2 illustrate the continuous format of facial valence and facial arousal features that were extracted from a person speaking truthfully and deceptively in 100-frame intervals of different videos in the dataset. A total of 2 facial affect features were extracted from each video frame.

*3.2.2 Visual* The OpenFace 2.2.0 toolkit [6] was used to extract visual features capturing facial action units (FAUs), eye gaze, and head pose from each facial frame of each video. The following OpenFace feature sets were computed: the intensities of 17 FAUs, eye gaze direction, and head pose. A total of 31 visual features were extracted from each video frame.

*3.2.3 Vocal* The OpenSMILE 2.0 toolkit [20, 41] was used to extract vocal features capturing cepstral, spectral, prosodic, and voice quality information from the audio associated with each video. The following three standard OpenSMILE feature sets were computed: the extended Geneva Minimalistic Acoustic Parameter Set (eGeMAPS)

[19], MFCC, and Prosody. eGeMAPS contains 23 features including frequency, energy, spectral, and cepstral information, which have been identified by experts for their potential to index affective and prosodic content in speech [19]. MFCC contains 39 features including 13 mel-frequency cepstral coefficients, and their first and second derivatives. Prosody contains 3 additional features related to fundamental frequency, loudness, and voicing. The eGeMAPs feature set contains some overlap with MFCC and Prosody sets. While we extracted all three sets, we did not include duplicate features in our feature selection and classification process. A total of 65 vocal features were extracted from each audio frame.

*3.2.4 Verbal* The Linguistic Inquiry and Word Count (LIWC) toolkit [23] was used to extract verbal features capturing psycholinguistic information from the transcript associated with each video. LIWC maps words in a transcript to a dictionary of 93 psychometric constructs and calculates the percentage of words in each category. Psychometric constructs include the attentional focus (e.g., pronouns, verb tense), emotions, social relationships (e.g., the speaker's level of dominance), and cognitive mechanisms (e.g., use of causal words) embedded in language. A total of 93 verbal features were extracted from each video transcript.

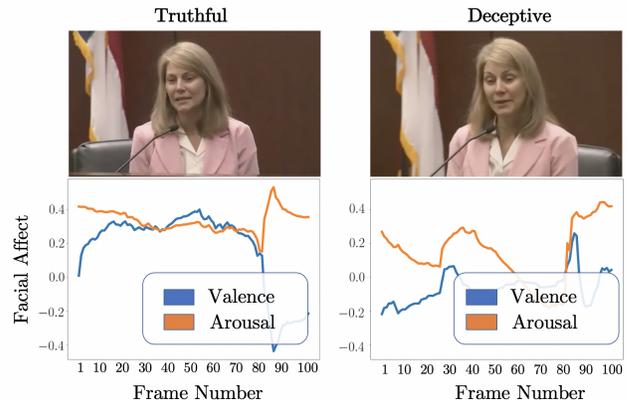

**Figure 2: Sample graphs of continuous facial valence and facial arousal during 100-frame intervals of a person speaking truthfully (left) and deceptively (right) in different videos.**

*3.2.5 Computing Fixed-Length Feature Vectors for Each Video* Features extracted from the facial affect, visual, and vocal modalities in each video are frame-by-frame and depend on the length of videos. In order to prepare these features for use in binary deception classification models, we represented each feature as a fixed-length vector of statistical attributes that captured the feature's temporal behavior and distribution during the variable-length videos, similar to [43, 51]. For each of the facial affect, visual, and vocal features, we used the TsFresh toolkit [11] to compute 130 time-series attributes, including the following statistical measures: mean, median, standard deviation, skew, kurtosis, maximum, minimum, sum of values, linear trends, autocorrelation with different lags, and the changes among values within different quantile ranges. Details regarding all of the computed functions are documented in the EfficientFCParameters class of TsFresh.

Features extracted from the verbal modality are contained in fixed 93-dimensional vectors for each video transcript, independent of video length. The final fixed-length feature vectors created with this process were of length 12833 representing each video with 260 facial affect features (2 x 130), 4030 visual features (31 x 130), 8450 vocal features (65 x 130), and 93 verbal features.

## 3.3 Feature Selection

In order to identify all relevant features for classification, we chose the Boruta algorithm [36] for feature selection; Boruta generates copies of original features with randomly mixed values, called shadow features, and then iteratively removes features that are less relevant ($\alpha < 0.05$) than their corresponding shadow features. We removed null and constant features across videos on training sets and performed feature selection through the Boruta algorithm (500 iterations). This process identified 30 facial affect features, 35 visual features, 38 vocal features, and 7 verbal features as useful for deception detection. The behavioral cues underlying these 110 selected features are summarized in Table 2.

## 3.4 Classification Experiments

Deception detection was formulated as a binary classification task to classify videos as truthful or deceptive. We conducted experiments with unimodal and multimodal linear SVM-based approaches to classify videos as truthful or deceptive, consistent with previous experiments with this dataset (summarized in Table 1).

*3.4.1 Experimental Setup* All training, hyperparameter tuning, and testing were conducted in the scikit-learn framework [46] with the Support Vector Classifier implementation of SVMs with a linear kernel. SVM hyperparameters C and $\gamma$ were each tuned on training sets with a grid-search in the discrete range [0.001, 0.01, 0.1, 1, 10, 100, 1000] during 5-fold stratified cross-validation. All features in the training and testing sets of each fold were standardized per the distribution of the training set, to prevent information leaking into the testing sets. Most of the 47 unique people in the dataset are only present in either deceptive or truthful videos. To account for this aspect of the dataset and avoid classification becoming a case of person-identification, we trained and tested models on distinct, randomly formed groups to ensure that the same person was not present in both the training and testing sets, consistent with previous experiments with this dataset [15, 51, 61]. All unimodal and multimodal classification experiments described below were conducted with 5-fold stratified cross-validation, split across 47 speaker identities, and repeated 10 times, maintaining the same proportion of truthful and deceptive speakers in each fold.

**Table 2: Behavioral Cues Underlying the Selected Features.**

| Modality | Behavioral Cues |
|---|---|
| Facial Affect | Valence |
| | Arousal |
| Visual [6] | Eye gaze direction: x, y, and z coordinates |
| | Eye gaze angle: x coordinates |
| | Head pose: y coordinates of location, rotation |
| | Facial Action Units: 1, 2, 5, 15, 17, 26, 45 |
| Vocal [20, 21] | Fundamental frequency ($F_0$) |
| | Voicing probability of $F_0$ |
| | MFCC: 2, 3, 4, 9, 12 |
| | MFCC first derivatives: 4, 10, 11 |
| | MFCC second derivatives: 0, 12 |
| Verbal [30] | Cognitive processes (e.g., use of causal words) |
| | Personal pronouns |
| | Differentiation (e.g., use of "hasn't" or "but") |
| | Numbers |
| | Negations |
| | Clout |
| | Authenticity |

*3.4.2 Evaluation Metrics* For each cross-validation fold, three metrics were computed: (1) AUC, area under the precision-recall curve, representing the probability of the classifier ranking a randomly chosen deceptive sample higher than a randomly chosen truthful one; (2) ACC, classification accuracy over the videos in the test set; (3) F1-score, weighted average of precision and recall. We computed the average AUC, ACC, and F1-score across all folds in order to evaluate each model's performance. Consistent with previous deception detection studies with this dataset, we used AUC as the primary metric for comparing the effectiveness of different approaches.

*3.4.3 Unimodal Models* We trained unimodal SVMs on the respective features of each modality to assess the effectiveness of individual modalities for deception detection.

*3.4.4 Early Fusion Models* We experimented with early fusion, which concatenates features from different modalities to form a single feature vector that is used in a classifier. Early fusion SVMs were trained on each of the 11 possible combinations of the 4 modalities. We used results from early fusion to assess the effectiveness of single classifiers in exploiting low-level interactions of features across modalities for deception detection.

*3.4.5 Non-Generative Ensembles* We experimented with 6 non-generative ensemble approaches [59], each with a different method of combining the decisions of unimodal classifiers to produce final predictions: (1) *hard and soft majority voting*, (2) *hard and soft stacking*, and (3) *hard and soft hybrid fusion*. In majority voting, the final classifier decision is either the class label predicted most frequently by the unimodal classifiers (hard majority voting) or the class label with the highest average predicted class probability across the unimodal classifiers (soft majority voting). In stacking, a final classifier is trained on either the predicted class labels of the unimodal classifiers (hard stacking) or the predicted class probabilities of the unimodal classifiers (soft stacking). In hybrid fusion, an early fusion vector with features from a set of modalities is concatenated with either the predicted class labels of the set of corresponding unimodal classifiers (hard hybrid fusion) or the predicted class probabilities of the set of corresponding unimodal classifiers (soft hybrid fusion), in order to create final feature vectors that are used in a classifier. All 6 approaches were trained on each of the 11 possible combinations of the 4 modalities. We used results from non-generative ensembles to assess the effectiveness of different approaches for combining the outputs of unimodal classifiers for deception detection.

*3.4.6 Generative Ensembles* We experimented with 2 generative ensemble approaches [59] that generate and train multiple classifiers on different subsets of multimodal input data: (1) *bagging* and (2) *boosting*. In bagging [32], random subsets of the training set are generated and used to train different classifiers, and the outputs of these models are combined into a final prediction by majority voting. Inspired by SVM boosting implemented on this dataset by [51] and for deception detection by [62], we experimented with AdaBoost (built from 50 estimators), with later classifiers focusing more on misclassified examples. The outputs of all classifiers are combined into a final prediction through weighted voting that assigns more accurate models a higher weight. Both generative ensemble approaches were trained on each of the 11 possible combinations of the 4 modalities. We used results from generative ensembles to assess the effectiveness of training multiple classifiers on different subsets of multimodal data for deception detection.

## 4 Analysis of Facial Affect and Deception

Since this paper introduces facial affect as a novel feature set in automated deception detection, this section analyzes the computational relationships observed between facial affect and deception in the high-stakes deception dataset used. We examined the distribution of facial affect across truthful and deceptive speakers, as well as the statistical significance of observed differences. A two-tail independent sample Welch's t-test [13] was used to determine significance levels of differences between the deceptive and truthful groups to account for differences in video lengths and not assume equal variances in feature distributions.

Across the videos, deceptive speakers exhibited significantly lower mean valence, median valence, and minimum valence than truthful speakers ($p<0.001$ for all) and significantly higher mean arousal, median arousal, and maximum arousal than truthful speakers ($p<0.002$ for all). Figure 3 graphs the probability density of the means of facial valence and facial arousal across all videos for deceptive and truthful speakers, computed with kernel density estimation (Gaussian kernel) [54]. The mean of facial valence for deceptive speakers was -0.07 and for truthful speakers was 0.06 (Figure 3). The mean of facial arousal for deceptive speakers was 0.21 and for truthful speakers was 0.13 (Figure 3).

Across the videos, deceptive speakers exhibited higher standard deviation in facial valence ($p<0.005$) and facial arousal ($p<0.002$) than truthful speakers. Figure 4 graphs the probability density of the standard deviation of facial valence and standard deviation of facial arousal for deceptive and truthful speakers, computed with kernel density estimation. Across videos, the mean standard deviation of facial valence for deceptive speakers was 0.14 and for truthful speakers was 0.11 (Figure 4). The mean standard deviation of facial arousal for deceptive speakers was 0.12 and for truthful speakers was 0.09 (Figure 4).

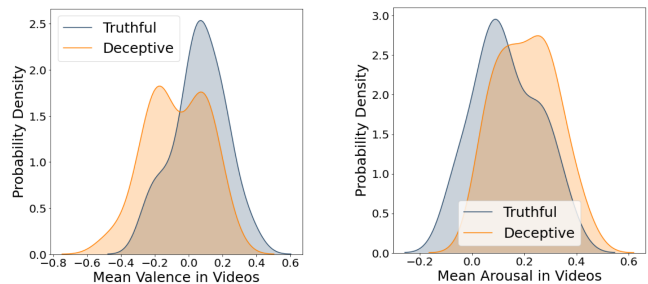

**Figure 3: Probability density distributions of the mean valence and mean arousal of deceptive (orange) and truthful (blue) speakers across all videos in the dataset.**

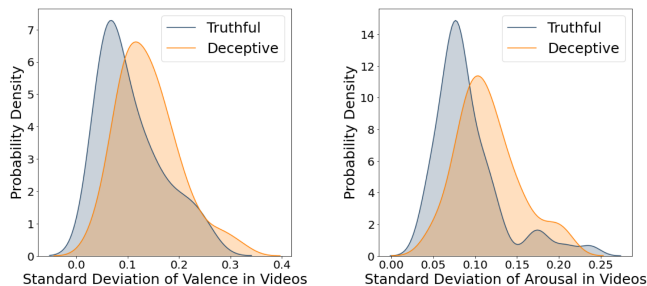

**Figure 4: Probability density distributions of the standard deviation of valence and the standard deviation of arousal of deceptive (orange) and truthful (blue) speakers across all videos in the dataset.**

The facial affect patterns seen in this dataset support the *leakage hypothesis* [18] and *four-factor theory* of deception [65], which proposed that deceivers, particularly those in high-stakes situations [50], experience and exhibit affective states with lower valence and higher arousal, possibly due to emotions such as anxiety, fear, and guilt. The high-stakes nature of the deception in this dataset may have contributed towards the significantly more negative and aroused affective states observed in deceptive speakers, given the serious consequences of communicating deceptively under oath in a courtroom. The significantly higher standard deviation of facial valence and facial arousal in the videos of deceptive speakers indicates that deceptive speakers exhibited more variation in affective states than truthful speakers in this dataset. *The significant differences observed in the facial valence and facial arousal of truthful and deceptive speakers in our research illustrate the discriminative potential of facial affect for detecting deception, making a case for using facial affect features in automated deception detection systems.*

## 5 Results and Discussion

Modeling results from the classification experiments discussed in Section 3.4 are presented in Table 3 and visualized in Figure 5. These results indicate the discriminative power of facial affect for automated deception detection in videos. Unimodal models trained on facial affect features achieved an AUC of 80% (Table 3). Facial affect contributed towards the highest-performing multimodal approach, which obtained an AUC of 91% through adaptive boosting (AdaBoost) across facial affect, visual, and vocal modalities (Table 3). This multimodal approach achieved an 84% accuracy, higher than the dataset benchmark (75% accuracy) [47] (Table 1). The 91% AUC achieved by our multimodal approach was also higher than the AUC of the best-performing automated approach on this dataset (88% AUC) (Table 1) that used interpretable visual, vocal, and verbal features with an SVM [61], but did not use affect.

Results from all unimodal classification experiments are discussed in Section 5.1. Results from the highest-performing early fusion, non-generative ensemble (soft hybrid fusion), and generative ensemble (AdaBoost) are discussed in Section 5.2. All statistical significance values of differences in model performance across classifiers and feature sets were computed with McNemar's chi-squared test ($\alpha$=0.05) with continuity correction, as described in [17]. Important features contributing towards the highest-performing model are discussed in Section 5.3 and listed in Table 4. Figure 5 visualizes the ROC curves of all 4 unimodal models and the best-performing multimodal models. A baseline model for deception detection (a classifier that always predicted "deceptive") would achieve 51% accuracy (55 deceptive videos out of 108 videos). We use this baseline when evaluating whether our models perform better than chance.

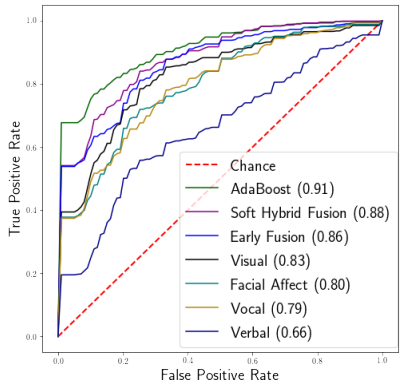

Figure 5: ROC curves for unimodal models and the best multimodal models from early fusion, non-generative ensemble (soft hybrid fusion), and generative ensemble (AdaBoost).

Table 3: Classification Results.

| Modality | AUC | ACC | F1 |
|---|---|---|---|
| **Unimodal Models** | | | |
| Facial Affect | 0.80 | 0.72 | 0.67 |
| Visual | 0.83 | 0.76 | 0.72 |
| Vocal | 0.79 | 0.72 | 0.68 |
| Verbal | 0.66 | 0.63 | 0.58 |
| **Early Fusion** | | | |
| Facial Affect + Visual | 0.86 | 0.74 | 0.71 |
| Facial Affect + Vocal | 0.85 | 0.76 | 0.73 |
| Facial Affect + Verbal | 0.81 | 0.76 | 0.72 |
| Visual + Vocal | 0.83 | 0.76 | 0.72 |
| Visual + Verbal | 0.85 | 0.73 | 0.65 |
| Vocal + Verbal | 0.77 | 0.67 | 0.63 |
| Facial Affect + Visual + Vocal | 0.86 | 0.79 | 0.78 |
| Facial Affect + Visual + Verbal | 0.83 | 0.68 | 0.54 |
| Facial Affect + Vocal + Verbal | 0.82 | 0.76 | 0.71 |
| Visual + Vocal + Verbal | 0.80 | 0.75 | 0.72 |
| Facial Affect + Visual + Vocal + Verbal | 0.85 | 0.78 | 0.75 |
| **Non-Generative Ensemble (Soft Hybrid Fusion)** | | | |
| Facial Affect + Visual | 0.83 | 0.74 | 0.72 |
| Facial Affect + Vocal | 0.83 | 0.74 | 0.70 |
| Facial Affect + Verbal | 0.77 | 0.69 | 0.64 |
| Visual + Vocal | 0.82 | 0.76 | 0.74 |
| Visual + Verbal | 0.78 | 0.71 | 0.68 |
| Vocal + Verbal | 0.77 | 0.67 | 0.63 |
| Facial Affect + Visual + Vocal | 0.88 | 0.81 | 0.80 |
| Facial Affect + Visual + Verbal | 0.79 | 0.72 | 0.70 |
| Facial Affect + Vocal + Verbal | 0.82 | 0.76 | 0.74 |
| Visual + Vocal + Verbal | 0.81 | 0.75 | 0.73 |
| Facial Affect + Visual + Vocal + Verbal | 0.85 | 0.80 | 0.78 |
| **Generative Ensemble (AdaBoost)** | | | |
| Facial Affect + Visual | 0.87 | 0.74 | 0.73 |
| Facial Affect + Vocal | 0.86 | 0.76 | 0.76 |
| Facial Affect + Verbal | 0.80 | 0.50 | 0.56 |
| Visual + Vocal | 0.87 | 0.78 | 0.78 |
| Visual + Verbal | 0.82 | 0.54 | 0.62 |
| Vocal + Verbal | 0.81 | 0.65 | 0.68 |
| **Facial Affect + Visual + Vocal** | **0.91** | **0.84** | **0.84** |
| Facial Affect + Visual + Verbal | 0.86 | 0.72 | 0.70 |
| Facial Affect + Vocal + Verbal | 0.86 | 0.78 | 0.77 |
| Visual + Vocal + Verbal | 0.86 | 0.78 | 0.77 |
| Facial Affect + Visual + Vocal + Verbal | 0.90 | 0.82 | 0.81 |

### 5.1 Performance of Unimodal Classifiers

Unimodal classification results revealed that unimodal models trained on facial affect, visual, and vocal features have significantly higher predictive power for deception detection, compared to those trained on the verbal modality, with p<0.001 for all pairwise comparisons. The AUCs achieved by unimodal facial affect, visual, and vocal classifiers were 80%, 83%, and 79%, respectively. Pairwise comparison revealed no significant differences in model predictions among these three modalities, demonstrating that a classifier trained on facial affect features, alone, is as discriminative as unimodal classifiers trained on visual and vocal modalities for deception detection. *These results support our hypothesis, suggesting that temporal patterns in facial affect can be used by machine learning models to effectively detect deception in high-stakes situations.* All unimodal models had a predictive power higher than chance, outperforming the baseline 51% accuracy. The low relative performance of the verbal modality indicates that the chosen verbal psychometric features (Table 2) were not as discriminative for detecting deception in this dataset, compared to other modalities, although it is worth noting that most of these cues (self-references, cognitive processes, negations) have been useful for detecting deception in other contexts [64, 65].

## 5.2 Performance of Multimodal Classifiers

As expected from past deception detection experiments [4, 10, 47, 51], unimodal models did not perform as well as multimodal approaches that have the advantage of integrating complementary and supplementary information across modalities [5] to detect the multimodal behavior of deception. The generative ensemble approach of AdaBoost with facial affect, visual, and vocal features outperformed other unimodal and multimodal approaches, as seen in Table 3. This approach attained an AUC of 91%, ACC of 84%, and F1-score of 84% and had significantly higher predictive power compared to boosting with the other 10 modality combinations ($p<0.02$ for all pairwise comparisons). Boosting was found to be more effective for deception detection in this dataset than feature-level and other decision-level multimodal fusion methods.

To better determine the contributions of each individual modality in our highest-performing multimodal approach, we conducted an ablation analysis. We examined the performance of the model, removing one modality at a time. Removing facial affect reduced model AUC by 4.1%. Removing visual reduced model AUC by 5.1%. Removing vocal reduced model AUC by 3.7%. *These results suggest that facial affect makes an effective contribution towards deception detection and merits consideration, along with visual and vocal features, when choosing modalities and feature sets to include in automated deception detection systems.*

## 5.3 Important Features

To identify important individual features that contributed towards the highest-performing multimodal approach, we examined the weights of each feature across the estimators. As noted in [9, 55], linear SVMs have an advantage of interpretability, because the trained model weights can be used for determining feature importance. The magnitude $[w_i]$ of each weight $w_i$ indicates the influence of the $i$th feature on the classifier's predictions. Table 5 lists the top 25 features used, which included 9 facial affect features, 2 visual features, and 14 vocal features, each representing statistical and time-series attributes of behavioral cues during speakers' communications. The 9 facial affect features identified as top contributors to the highest-performing multimodal model included the autocorrelation of arousal, properties of the distribution of arousal (e.g., standard deviation), and properties of the distribution of valence (e.g., kurtosis). The 2 visual features and 14 vocal features identified as top contributors to the highest-performing multimodal model included patterns in FAUs (inner brow raising and blinking), fundamental frequency, and MFCC coefficients. *These results suggest that temporal patterns in facial affect features can be used in multimodal*

Table 4: Top 25 features used by the highest-performing multimodal model, per SVM feature weights.

| Modality | Feature |
|---|---|
| Facial Affect | Arousal: agg autocorrelation (lag 40) |
| | Arousal: CIQ [†] [0.4,0.6], [0.6,1] |
| | Arousal: longest strike below mean |
| | Arousal: standard deviation |
| | Valence: kurtosis |
| | Valence: CIQ [0, 0.1], [0,1] |
| | Valence: sum of values |
| Visual | FAU 1: partial autocorrelaton (lag 6) |
| | FAU 45: longest strike below mean |
| Vocal | F0: CIQ [0, 0.9], [0, 1] |
| | MFCC 0 2nd deriv: mean change |
| | MFCC 4: kurtosis |
| | MFCC 4 1st deriv: CIQ [0.6, 1] |
| | MFCC 9: partial autocorrelation (lag 6) |
| | MFCC 10 1st deriv: median, CIQ [0, 0.6] |
| | MFCC 10 2nd deriv: mean change |
| | MFCC 11 1st deriv: CIQ [0.2, 0.4] |
| | MFCC 12: CIQ [0, 0.1], [0.2, 0.4], [0.2, 1], [0.4, 1] |

[†] CIQ = change in quantiles

*models, along with visual and vocal features, to effectively detect deception in high-stakes situations.*

## 6 Conclusion

This paper presents a novel analysis of the discriminative power of facial affect for automated deception detection, along with interpretable features from visual, vocal, and verbal modalities, in real-world high-stakes courtroom situations in-the-wild. Facial affect contributed towards the highest-performing multimodal approach, which achieved an AUC of 91% through adaptive boosting trained on facial affect, visual, and vocal features. Our results indicate that temporal patterns in facial valence and facial arousal have potential as features in automated deception detection. We found significant differences in distributions of facial valence and facial arousal between truthful and deceptive speakers, contributing computational support to the *leakage hypothesis* and *four-factor theory* from psychological research on relationships between affective states in deceptive behavior.

This paper demonstrates the potential for developing automated deception detection approaches that leverage deep neural networks trained on large emotion datasets collected in-the-wild (in our case, AffWildNet trained on the Aff-Wild dataset) to extract facial valence and facial arousal as features for deception detection. Our research provides a proof-of-concept and motivation for future work towards developing affect-aware systems for automatically detecting deception and other social behaviors, particularly those occurring in unconstrained situations in-the-wild.

## Acknowledgments

This research was supported by the USC Provost's Undergraduate Research Fellowship from the Office of the Provost at the University of Southern California.